\documentclass[10pt,twocolumn,letterpaper]{article}

\usepackage[pagenumbers]{cvpr} 

\definecolor{cvprblue}{rgb}{0.21,0.49,0.74}
\usepackage[pagebackref,breaklinks,colorlinks,allcolors=cvprblue]{hyperref}

\usepackage{graphicx}
\usepackage{wrapfig}
\usepackage{multirow}
\usepackage{float}
\usepackage{amsmath}
\usepackage{tabu}
\usepackage{multicol}
\usepackage{makecell}
\usepackage{booktabs}
\usepackage{nameref}

\usepackage{xcolor}
\usepackage{colortbl}
\definecolor{color3}{RGB}{255, 255, 185}
\definecolor{color2}{RGB}{255, 216, 178}
\definecolor{color1}{RGB}{255, 178, 178}
\newcommand{\cc}[1]{\cellcolor{color#1}}

\def\paperID{*****} 
\def\confName{CVPR}
\def\confYear{2026}

\title{MOFA-VTON: More Fashion Possibilities with Fine-Grained Adaptations in Virtual Try-On}
\author{Xiaoyu Han$^{*,1}$, Chenyang Wang$^{*,1}$, Jing Wang$^2$, Shunyuan Zheng$^1$, \\ Quanling Meng$^1$, Shengping Zhang$^{\dag,1,3}$\\
$^1$Harbin Institute of Technology, $^2$ HiDream.ai Inc., \\ 
$^3$Harbin Institute of Technology (Weihai) Qingdao Research Institute\\
\tt\small xyhan@stu.hit.edu.cn, c.wang@hit.edu.cn, jwang@njust.edu.cn,\\  
\tt\small sawyer0503@hit.edu.cn,  quanling.meng@hit.edu.cn, s.zhang@hit.edu.cn}

\begin{document}

\twocolumn[{
\maketitle
\begin{figure}[H]
\hsize=\textwidth
\centering
\includegraphics[width=\textwidth]{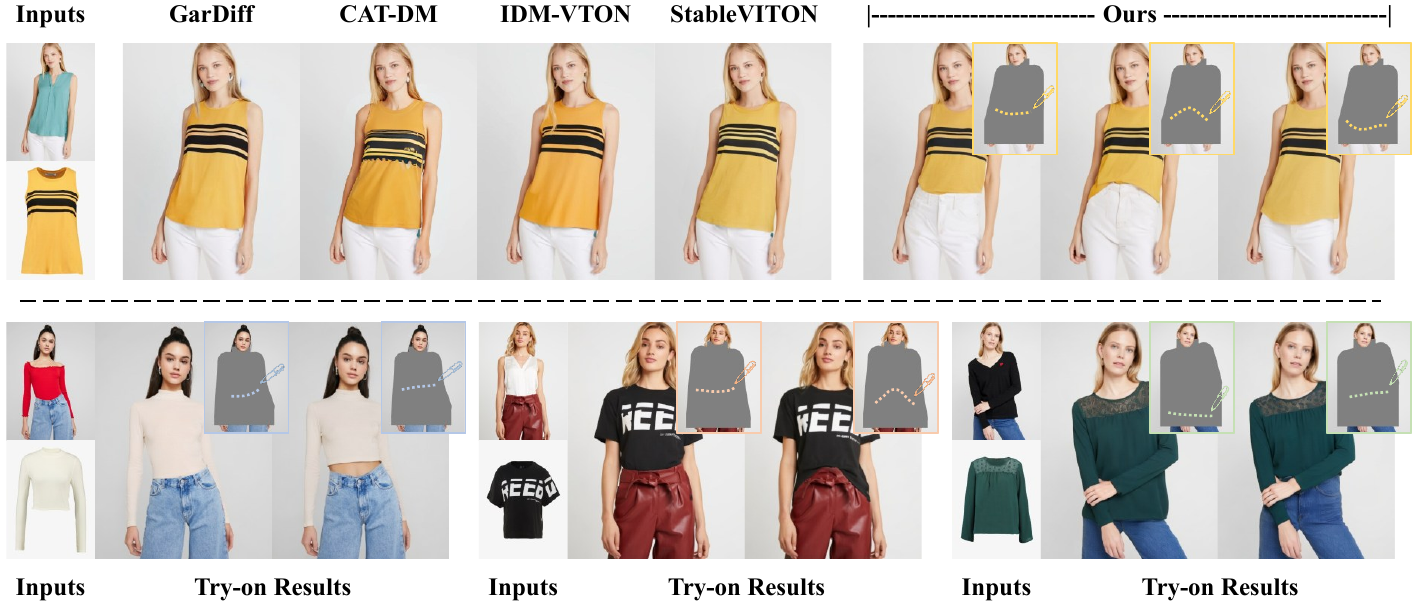}
\caption{\textbf{Example try-on results from existing methods and our MOFA-VTON.} The first row shows a comparison and the second row presents our more results. By allowing users to control the interaction between upper and lower clothing with sketches, MOFA-VTON breaks free from the limitations of traditional virtual try-on with a single fixed result, unlocking more possibilities for virtual fashion try-on.}
\label{fig:teaser}
\end{figure}
}]

\let\thefootnote\relax\footnotetext{$^*$ Equal contribution}
\let\thefootnote\relax\footnotetext{$^\dag$ Corresponding author (s.zhang@hit.edu.cn)}

\begin{abstract}
Virtual try-on aims to fit an in-shop clothing image onto a specific human body. An optimal virtual try-on method should provide diverse and flexible dressing options, accurately reflecting the varied wearing styles encountered in real-life scenarios, tailored to individual preferences and fashion aspirations. However, current methods predominantly perform a direct replacement of the original clothing with the target clothing, following the same dressing pattern. This limited control over clothing adaptation may result in fixed and monotonous try-on outputs. To delve into \textbf{Mo}re Fashion Possibilities with \textbf{F}ine-Grained \textbf{A}daptations in Virtual Try-On, we propose a novel virtual try-on method, termed \textbf{MOFA-VTON}, which allows adjustment for clothing adaptations in try-on results through simple sketches by users. Specifically, we first design a mask construction strategy that transforms user-drawn curve sketches into a dual-region mask, replacing the traditional clothing-agnostic mask and providing fine-grained layout guidance for the subsequent generation process. Further, we propose layout adjustment blocks that utilize the cross-attention mechanism to independently learn layout correspondences for upper and lower regions of the human body, refining the spatial arrangement of the two regions. With these implementations, our method enables flexible and fine-grained adaptations of target clothing, overcoming the constraints of a fixed layout. Extensive experiments on VITON-HD and DressCode datasets demonstrate that our proposed MOFA-VTON outperforms previous state-of-the-art methods and provides more fashion possibilities for virtual try-on.
\end{abstract}
  
\section{Introduction}
In recent years, the e-commerce industry has developed rapidly, leading to an increasing number of consumers purchasing clothing online. 
To offer an experience comparable to in-store fitting, virtual try-on technology~\cite{viton, cp-vton, clothflow, pf-afn, stableviton} has emerged, which aims to fit the target clothing onto a specific body to simulate a realistic try-on effect.

Early methods~\cite{viton-hd, cp-vton+, viton-gt, fashionmirror, cit} are typically based on conventional generative networks, such as Generative Adversarial Networks (GANs)~\cite{gan}. Then, with the rapid development of diffusion models~\cite{ddpm, ldm}, more recent methods~\cite{MGD, scw-vton, tryondiffusion, stableviton, kgi} leverage their powerful generative capabilities to achieve higher-quality virtual try-on.
Despite achieving advancements, previous methods overlook a crucial aspect of the try-on process, namely, the varied fashion possibilities.
Specifically, most existing methods focus solely on overlaying clothing onto the body, merely replicating the adaptation of the original clothing.
As illustrated in Figure~\ref{fig:teaser}, the results of all baseline methods constrain the lower boundary of the clothing to the same position.
However, in reality, individuals have personal preferences and fashion pursuits. Different persons may wear the same clothing differently, and even the same person may have different dressing ideas due to different occasions. 
For example, tucking in clothing creates a sharp look, while leaving them untucked gives a casual style (see Figure~\ref{fig:teaser}, first row, right side).
Therefore, an ideal virtual try-on method should not only generate high-quality results but also offer greater flexibility in clothing adaptation, enabling personalized adjustments that allow users to explore various styling possibilities beyond simple overlaying.
However, most existing methods struggle to achieve the desired effect. 
This can be mainly attributed to two factors: (1) the conventional clothing-agnostic mask~\cite{viton-hd} limits the adaptability of the target clothing; (2) the inability to dynamically adjust clothing features hinders the generation of diverse try-on results.

To address the above problems, we propose MOFA-VTON, a novel virtual try-on method, to explore more fashion possibilities with diverse adaptations in virtual try-on. 
As illustrated in Figure~\ref{fig:teaser}, MOFA-VTON enables users to control the clothing adaptation via user-drawn curve sketches, achieving diverse try-on effects while maintaining high-quality generation.
Specifically, we first propose a dual-region mask to replace the clothing-agnostic mask used by most previous methods, which is transformed from the user-drawn sketch using a mask construction strategy based on the DensePose map~\cite{densepose}.
With the boundary defined by the curve sketch, the mask can separately represent the upper and lower regions of the human body, which provides explicit and fine-grained layout guidance for the subsequent generation process.
Then, we introduce a denoising UNet~\cite{unet} as the base network of try-on generation, termed Adapt-Net. Building on this, we employ a CLIP~\cite{clip} image encoder, a pre-trained Cloth-Net model, and a novel region encoder to extract clothing features at different levels and inject them into the Adapt-Net. 
To further achieve customized try-on effects, we propose layout adjustment (LA) blocks, which are integrated into the intermediate layers and decoder layers of Adapt-Net.
The layout adjustment blocks independently learn the layout correspondence for two regions through a cross-attention mechanism, while additionally applying transposed convolution and mask filtering to refine clothing region features to specific positions.

Our contributions could be summarized as follows:
\begin{itemize}
    \item We propose a novel user-interactive virtual try-on method, MOFA-VTON, which achieves more fashionable and diverse clothing adaptations while maintaining high-quality try-on results.
    \item We design a dual-region mask, derived from user-drawn curve sketches using a mask construction strategy, to provide explicit and fine-grained layout guidance for the subsequent try-on generation.
    \item We propose layout adjustment blocks that refine the spatial layout of the clothing at the feature level, ensuring the clothing adaptation accurately matches the user's sketch.
    \item Extensive experiments on the VITON-HD and DressCode datasets demonstrate the state-of-the-art performance and fine-grained controllability of our MOFA-VTON compared to existing virtual try-on methods.
\end{itemize}
\section{Related Work}
\label{sec:work}

\noindent\textbf{Image-based Virtual Try-on.}
Early methods~\cite{viton, z-flow, sc-vton, sal-vton} for virtual try-on are most built upon conventional generative network, which generally consists of two critical stages: clothing warping and try-on synthesis.
First, for the clothing warping stage, VITON~\cite{viton} employs the thin-plate spline (TPS) transformation~\cite{TPS} to align target clothing with the human body, followed by an encoder-decoder framework to generate the final try-on result.
To address the insufficient degrees of freedom in TPS, ClothFlow~\cite{clothflow} introduces appearance flow~\cite{appearance_flow}, enabling pixel-level displacement for more precise warping.
Second, there are also some methods~\cite{vtnfp, acgpn, pl-vton, pl-vton2, outfit-viton} that focus on improving the synthesis stage.
For example, HR-VTON~\cite{hr-vton} designs a framework that integrates clothing warping and try-on synthesis into a unified process, effectively reducing misalignment issues and preventing pixel squeezing artifacts.
However, many early methods for virtual try-on are prone to unstable training and may cause artifacts or unnatural details in fine-grained regions, making it a significant challenge to generate robust and high-quality results.

In recent years, leveraging the powerful generative ability of the diffusion model~\cite{ddpm, ldm}, more related studies~\cite{d4-vton, mm_vto, anyfit, boow-vton, leffa, ita-mdt} adopt it as a core component to achieve more realistic virtual try-on.
For instance, StableVITON~\cite{stableviton} utilizes zero cross-attention blocks within the diffusion decoder to learn semantic relationships between clothing and the person, achieving better detail preservation. 
CAT-DM~\cite{cat-dm} introduces a truncation-based acceleration strategy, improving the controllability of diffusion models and speeding up the sampling process. 
IDM-VTON~\cite{idm-vton} employs two parallel UNets along with IP-adapter~\cite{ip-adapter} and detailed natural language clothing descriptions, effectively enhancing the clothing fidelity.

\begin{figure*}
    \centering
    \includegraphics[width=1\linewidth]{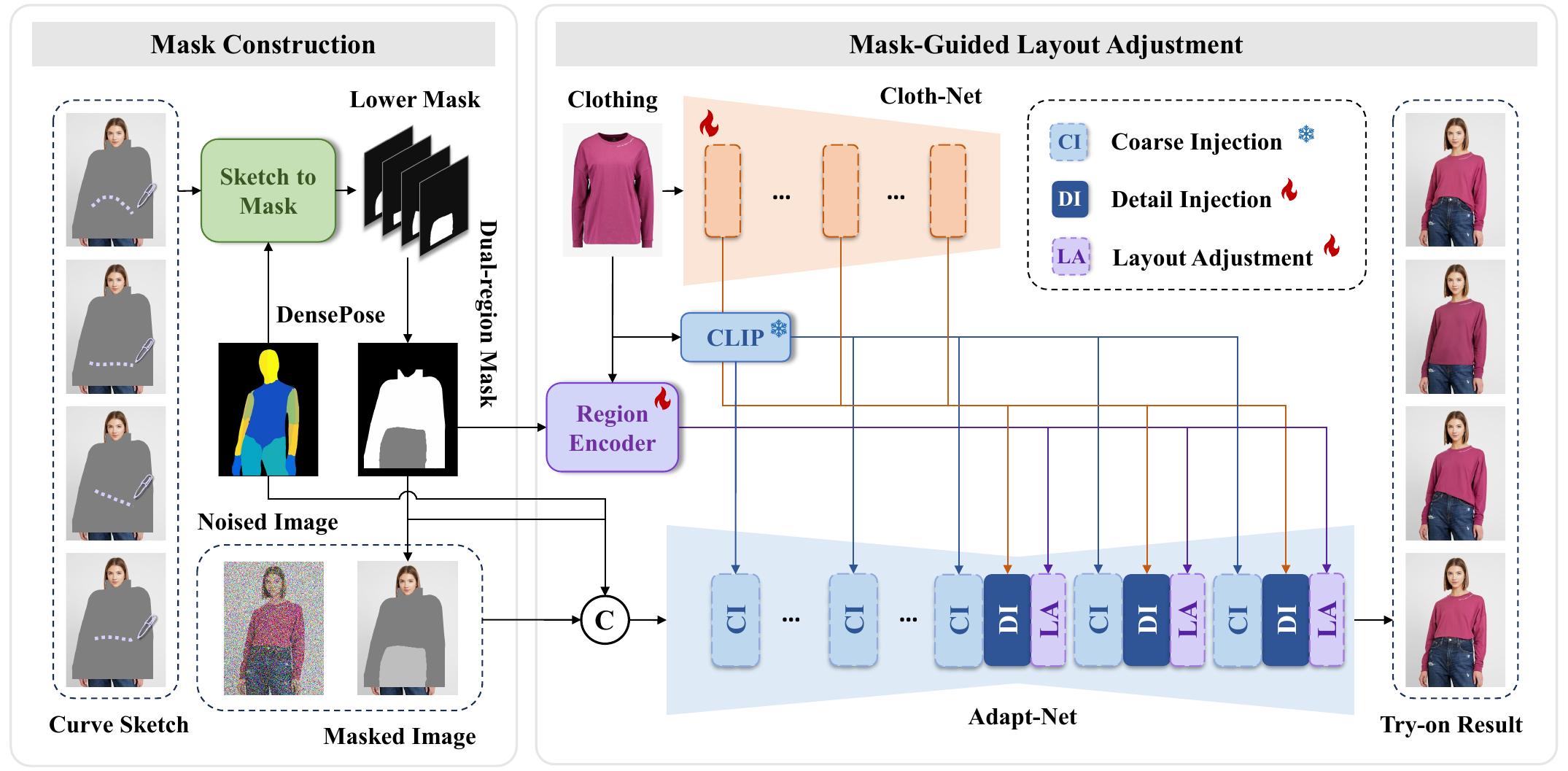}
    \caption{\textbf{Overview of MOFA-VTON.} Given a user-drawn curve sketch, we derive the corresponding dual-region mask and the masked person image through a mask construction strategy, both of which are fed into the Adapt-Net. Additionally, clothing features at multiple levels are extracted from CLIP, Cloth-Net, and a region encoder, and are then injected into the Adapt-Net to preserve the clothing appearance. To match the layout of these features with the drawn curve, layout adjustment blocks are incorporated within the Adapt-Net to refine the spatial arrangement of the upper and lower regions.}
    \label{fig:overview}
\end{figure*}

\noindent\textbf{Virtual Try-on for Diversity.}
In the virtual try-on task, although high-quality generation is crucial, it is equally important to consider the diverse try-on effects that reflect the various dressing styles in real life.
However, this capability for diversity remains largely unexplored, with only a few methods~\cite{controlling, cotton, wear-any-way, promptdresser, up-vton} exploring possible solutions.
COTTON~\cite{cotton} introduces a landmark-guided transformation for clothing warping, enabling vertical adjustments of clothing based on predicted landmarks and segmentation.
Wear-Any-Way~\cite{wear-any-way} proposes a sparse alignment strategy, guiding clothing generation through sparse point-based control at specific locations. 
A common characteristic of these methods is that they are all point-based and require manual adjustment of each control point to achieve diverse effects. 
However, the influence region of each discrete point is vague, and the limited degrees of freedom restrict fine-grained control over clothing details.
In addition, PromptDresser~\cite{promptdresser} and UP-VTON~\cite{up-vton} propose text-driven methods that allow altering clothing via natural language instructions.
However, such text-based control still struggles to achieve pixel-level adjustments and may introduce semantic deviations, resulting in mismatches between generated and input clothing appearances.
In contrast, we propose using the curve sketch to construct a dual-region mask and perform spatial layout adjustments at the pixel level, thereby enabling more precise and diverse try-on effects.
\section{Proposed Method}
\subsection{Preliminary}
\noindent\textbf{Stable Diffusion.}
Stable Diffusion~\cite{ldm} is a deep generative model that synthesizes high-quality images conditioned on auxiliary information. 
In the forward diffusion process, the input image $x$ is first transformed to latent features by a pre-trained autoencoder: $z_0=\mathcal{E}(x)$.
Then $z_0$ is gradually added Gaussian noise over $T$ time steps, producing a sequence \( \{z_t\}_{t=1}^{T} \). This process is modeled as:  
\begin{equation}
z_t = \sqrt{\alpha_t}z_{t-1} + \sqrt{1-\alpha_t}\epsilon,
\end{equation}
where \( \epsilon \sim \mathcal{N}(0, 1) \) is the added noise, \( \alpha_t \) is a noise schedule parameter controlling the noise level at each step. 
In the reverse process, a denoising network \( \epsilon_\theta \) recovers the original image $z_0$ by predicting the added noise in the latent space, which is trained over multiple time steps to minimize: 
\begin{equation}
\mathcal{L}_{\text{ldm}} = \mathbb{E}_{z_0, \epsilon, t} \left[ \|\epsilon - \epsilon_\theta(z_t, c, t) \|^2_2 \right],
\end{equation}
where \( c \) is the additional condition.  

\subsection{MOFA-VTON Pipeline}
Given a person image, a user-drawn curve sketch, a DensePose map, and an in-shop clothing image, the key challenge in more fashion possibilities try-on is obtaining flexible dressing layout guidance and dynamically adjusting the features of the corresponding regions based on the guidance.
As shown in Figure~\ref{fig:overview}, our proposed MOFA-VTON achieves this via two steps: (1) dual-region mask construction derived from user-drawn curve sketches, and (2) mask-guided layout adjustment.
Specifically, MOFA-VTON first constructs a dual-region mask according to the desired curve sketch, which provides layout guidance for the following generation process (Section~\ref{sec:DMC}).
Next, a diffusion-based model, Adapt-Net, is proposed as the foundational architecture for the generation process. It takes as input the concatenation of a noised image $\mathcal{E}(I)$, a DensePose map $\mathcal{E}(P)$, a masked image $\mathcal{E}(I_m)$, and a dual-region mask $M_d$, and then performs denoising on $\mathcal{E}(I)$. 
To integrate the appearance of the target clothing, the CLIP encoder, Cloth-Net, and the region encoder extract features from the in-shop clothing image $C$ at different levels and guide the denoising process via elaborately designed blocks within Adapt-Net (Section~\ref{sec:MLA}).
These modules ensure a comprehensive and contextually rich encoding process, crucial for high-fidelity and flexible try-on results.

\subsection{Dual-region Mask Construction}
\label{sec:DMC}
\begin{figure}
    \centering
    \includegraphics[width=1\linewidth]{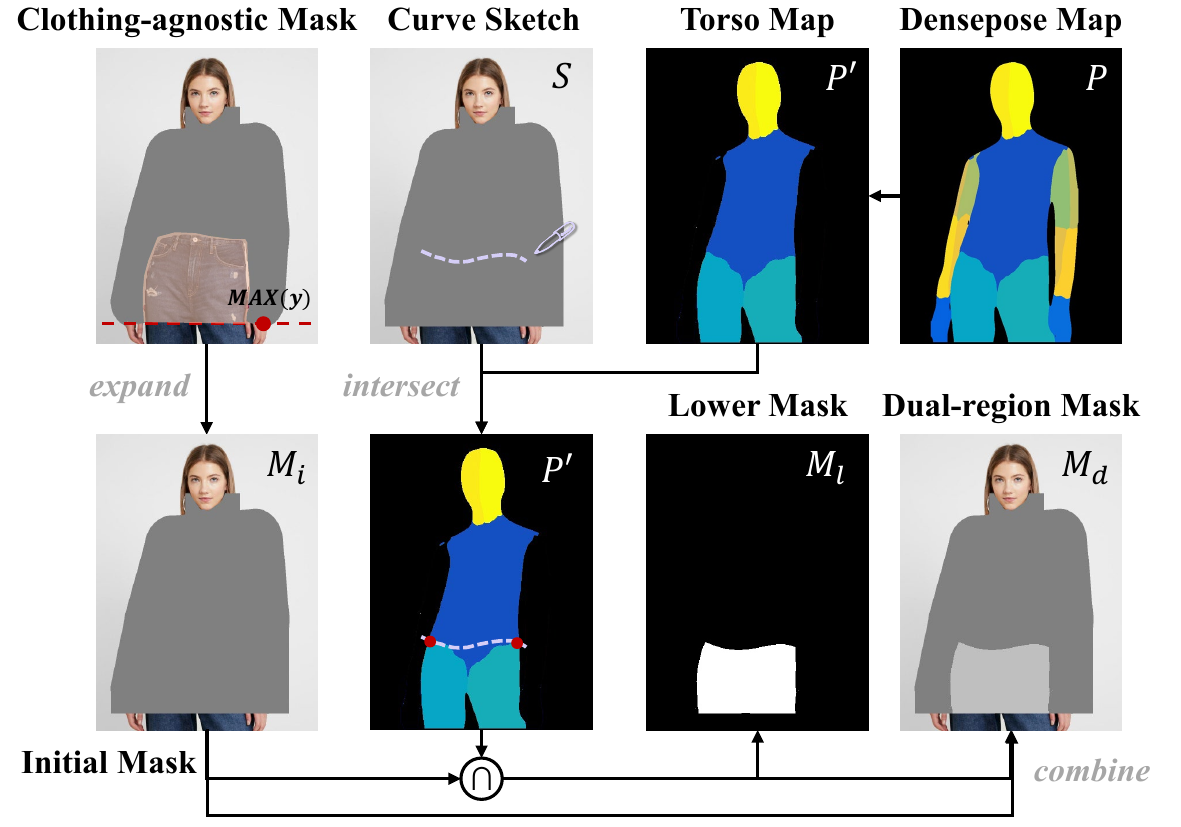}
    \caption{\textbf{Procedure of the mask construction strategy.} It converts a user-drawn curve sketch into a dual-region mask. For clarity, some masks in the figure are displayed on the person image.}
    \label{fig:mask}
\end{figure}

Most existing methods adopt a clothing-agnostic mask to remove the original clothing appearance while retaining other regions, which enables self-supervised training.
However, taking the upper-body try-on as an example, fully exposed lower regions (e.g., pants or skirts) restrict new clothing to only be generated in the same layout as the original clothing, resulting in a fixed and monotonic try-on effect.
In contrast, MOFA-VTON supports a more flexible try-on process. Users can achieve the desired try-on effect by simply sketching the corresponding curves.
To convert the curve sketch into a region representation for layout guidance, we design a mask construction strategy to generate a dual-region mask.
Specifically, as shown in Figure~\ref{fig:mask}, we first modify the conventional clothing-agnostic mask by expanding it based on its lowest point to obtain the initial mask $M_i$, which additionally covers the interaction between the upper and lower regions.
Then, we use the DensePose map $P$ as the basic body silhouette, as it precisely captures the body contours without interference from clothing. 
We remove the arm region from the DensePose map $P$ and get a torso map $P'$, then truncate the user-drawn curve at its intersection points with the torso region. 
Subsequently, we sample points along the truncated curve and refit them to ensure a smoother shape while eliminating unreasonable points. 
Note that if the original curve does not reach the required length, we extend it to the torso boundary according to the fitted results.
Next, we use the refined curve segment to crop the torso map $P'$ and extract the region that only contains the lower body. After binarization and intersection with the initial mask $M_i$, we obtain the mask representing the lower region, which is defined as the lower mask $M_l$. 
Finally, by combining $M_l$ with the initial mask $M_i$, we are able to derive the final dual-region mask $M_d$:
\begin{equation}
    M_d = \beta M_{i} * (1-M_{l}) + \gamma M_{l},
\label{eq:mask}
\end{equation}
where $\beta$ and $\gamma$ are defined as the combination weights assigned to the upper and lower regions, respectively, which are set to 1 and 0.5 in this paper.

\begin{figure*}
    \centering
    \includegraphics[width=1\linewidth]{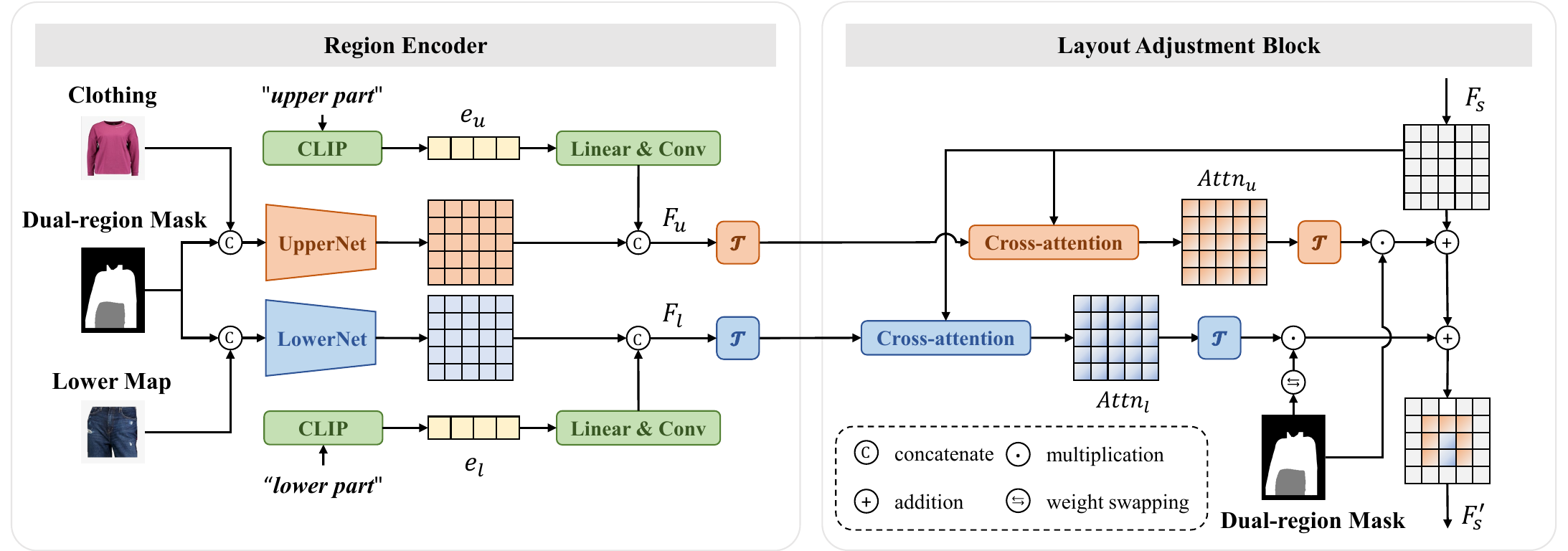}
    \caption{\textbf{Schematic of the region encoder and layout adjustment block.} The region encoder extracts upper and lower region features $F_{u}$ and $F_{l}$, which are then processed in the layout adjustment block to learn the layout correspondence with the feature $F_s$ from Adapt-Net.}
    \label{fig:module}
\end{figure*}

\subsection{Mask-guided Layout Adjustment}
\label{sec:MLA}
We employ a denoising UNet as the backbone of our network for generating try-on results, termed Adapt-Net. 
To enhance the representation of target clothing, a CLIP image encoder and Cloth-Net are integrated to extract detailed features of in-shop clothing, which are then injected into the Adapt-Net.
Specifically, in Adapt-Net, we introduce coarse injection (CI) blocks and detail injection (DI) blocks, which are implemented based on the design in ControlNet~\cite{controlnet}.
These blocks use the features from the Adapt-Net backbone as the query to interact with the features extracted by the CLIP image encoder and Cloth-Net, respectively, thereby injecting both high-level semantic features and fine-grained semantic details into Adapt-Net.
However, relying solely on the above components, the extracted clothing features lack clear layout guidance, leading to generated clothing failing to interact properly with other regions.
To address this problem, we adopt a region encoder and a set of layout adjustment blocks. 
The former further extracts texture features from the in-shop clothing image and the lower body of the person image separately and feeds them into the latter, which guides different features to the specified regions while also restoring the content occluded due to mask expansion in the construction strategy.

\noindent\textbf{Region Encoder.}
As shown in Figure~\ref{fig:module}, the region encoder adopts a symmetric architecture. 
Regarding the upper region, we first use a network consisting of stacked convolution layers (referred to as UpperNet) to encode the inputs of in-shop clothing $C$ and the dual-region mask $M_d$, then we contact its output with the corresponding text embeddings $e_u$, which are extracted by the CLIP text encoder with the predefined prompt. 
Finally, we obtain the upper region features $F_u$. The above process can be represented as:
\begin{equation}
F_u = Concat(E_u(Concat(C, M_d), \phi(p_u))),
\end{equation}
where $E_u$ denotes the UpperNet, $p_u$ is the text prompt for the upper region, and $\phi$ represents the CLIP text encoder with the subsequent convolutional and linear layers.
Similarly, the lower region features $F_l$ is obtained as follows:
\begin{equation}
F_l = Concat(E_l(Concat(B, M_d), \phi(p_l))),
\end{equation}
where $E_l$ denotes the LowerNet, and $p_l$ is the predefined prompt that matches the lower region. The lower map $B$ is obtained from the person image $I$ via human parsing~\cite{human_parsing}, which describes the content of the lower body. 
Before outputting the region features, we introduce a transposed convolution layer $\mathcal{T}$ at the final stage. The purpose is to leverage the upscaling capability of transposed convolution to propagate region features to the surroundings, expanding their influence. Meanwhile, it can encourage the model to adaptively learn feature interactions at the boundaries of the two regions.
Therefore, the final outputs of the region encoder are $\mathcal{T}(F_u)$ and $\mathcal{T}(F_l)$. 

\begin{figure*}[t!]
    \centering
    \includegraphics[width=1\linewidth]{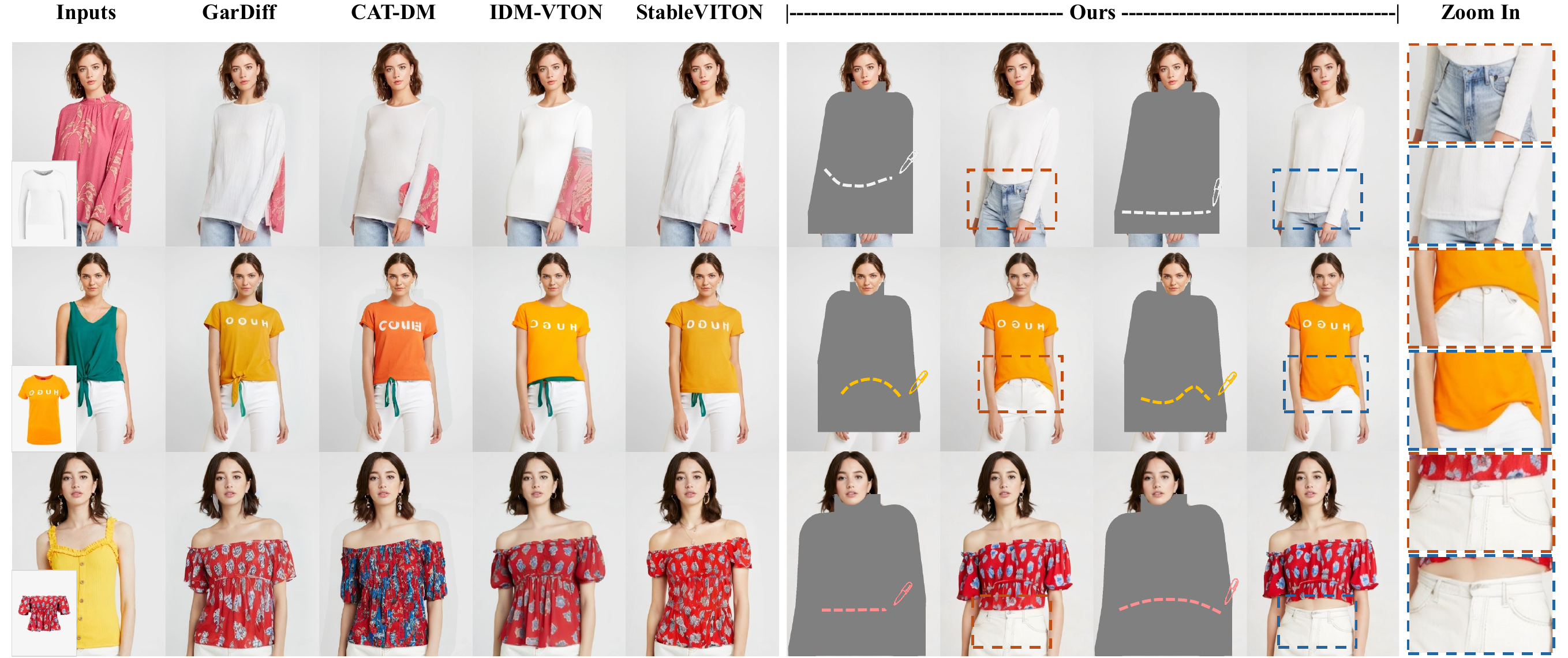}
    \caption{\textbf{Qualitative comparison with baseline methods on the VITON-HD dataset.} Columns 6–9 show the results corresponding to different curves generated by our method, while the last column presents a zoomed-in view.}
    \label{fig:quality}
\end{figure*}

\noindent\textbf{Layout Adjustment Block.} 
We design the layout adjustment block to condition the region features $\mathcal{T}(F_u)$ and $\mathcal{T}(F_l)$ into Adapt-Net, adjusting the features of different regions to their corresponding spatial locations.
As shown in Figure~\ref{fig:module}, the layout adjustment block mainly contains two cross-attention components, where the feature map $F_s$ from the detail injection (DI) block is fed as the query, while the key and value are from the region features $\mathcal{T}(F_u)$ or $\mathcal{T}(F_l)$. 
The learned correspondences enable the region information in $F_u$ and $F_l$ to be effectively injected into the features extracted from earlier layers of Adapt-Net, thereby facilitating layout adjustment. The process is presented below:
\begin{equation}
Attn_u = softmax(\frac{F_sW_q \cdot (\mathcal{T}(F_u)W_{ku})^T}{\sqrt{d_k}}) \cdot \mathcal{T}(F_u)W_{vu},
\end{equation}
\begin{equation}
Attn_l = softmax(\frac{F_sW_q \cdot (\mathcal{T}(F_l)W_{kl})^T}{\sqrt{d_k}}) \cdot \mathcal{T}(F_l)W_{vl},
\end{equation}
where $W_q$, $W_{ku}$, $W_{vu}$, $W_{kl}$, $W_{vl}$ are both linear layers within the attention mechanism, and $d_k$ is the dimension of the key. 
Similar to the region encoder, there is also a transposed convolution layer $\mathcal{T}$ after the cross-attention component, obtaining $\mathcal{T}(Attn_u)$ and $\mathcal{T}(Attn_l)$.
To further enhance the region-specific adaptation, we perform a mask filtration operation using the dual-region mask $M_d$ after $\mathcal{T}$, constraining the propagated features within their intended regions and preventing unintended blending with others.
With different weight ratios assigned to the upper and lower regions, the dual-region mask $M_d$ effectively adjusts feature mixing at the boundary, ensuring a smoother transition and facilitating more precise region interactions.
Overall, the above process can be formulated as:
\begin{equation}
F’_s = F_s + \mathcal{T}(Attn_u) \cdot M + \mathcal{T}(Attn_l) \cdot M',
\end{equation}
where $M'$ is obtained by swapping the weights in the dual-region mask $M$: $M'=\beta+\gamma-M$, with $\beta$ and $\gamma$ defined in Equation~\ref{eq:mask}.
$F’_s$ is progressively optimized through each layer of Adapt-Net, ultimately producing the final try-on result decoded from the latent space.
\section{Experiment}

\begin{figure}
    \centering
    \includegraphics[width=0.95\linewidth]{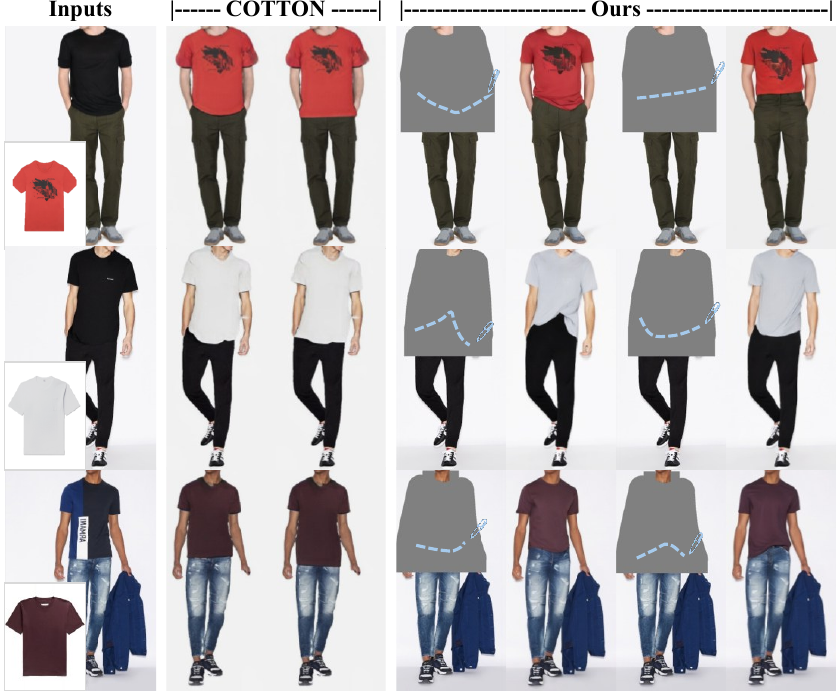}
    \caption{\textbf{Qualitative results on the DressCode dataset.}}
    \label{fig:quality2}
\end{figure}

\subsection{Datasets}
We conduct main experiments on the public virtual try-on benchmark dataset VITON-HD~\cite{viton-hd}, which provides 13,679 pairs of person images and in-shop clothing images, along with corresponding human representations such as DensePose maps and human parsing results.
To further validate the effectiveness of our method, we perform additional experiments on the DressCode~\cite{dresscode} dataset, which is also a widely used dataset that includes a broader range of cases.

\begin{figure*}
    \centering
    \includegraphics[width=1\linewidth]{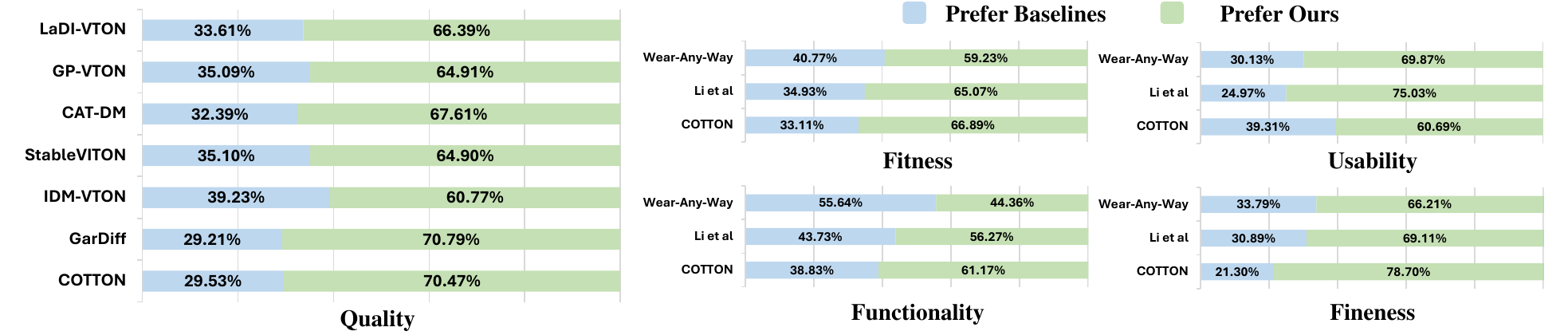}
    \caption{\textbf{User study of baseline methods and our method.} MOFA-VTON is compared against each baseline in an A/B evaluation.}
    \label{fig:user_study}
\end{figure*}

\subsection{Qualitative Results}
Figure~\ref{fig:quality} presents qualitative results on the VITON-HD dataset, where we compare our MOFA-VTON with several open-source baseline methods.
We evaluate these results in terms of both visual quality and diversity.
Firstly, compared to existing methods, MOFA-VTON exhibits greater robustness and effectively avoids interference from irrelevant content in the original image, producing more plausible and higher-quality virtual try-on results. 
For example, in the second case of Figure~\ref{fig:quality}, all compared methods fail to remove the ribbon from the original image, resulting in unnatural remnants in the outputs. 
In contrast, MOFA-VTON successfully handles such complex cases, achieving more visually coherent clothing adaptation.
On the other hand, most existing methods struggle to achieve diverse try-on effects, as they rigidly adhere to the original clothing adaptation, producing results that closely mirror the input image and lack variation in wearing styles.
Especially for the third example, it is difficult for them to properly adapt to the shorter target clothing and instead forcibly stretch it, which disrupts the natural proportions.
Conversely, MOFA-VTON supports more flexible layout adjustments, achieving various effects while staying within a reasonable range.
Overall, as shown in Figure~\ref{fig:quality}, MOFA-VTON not only achieves both tucked and untucked effects but also enables diverse hemline variations through fine-grained control, such as curved shapes and the waist-revealing effect.

Further qualitative experiments on the DressCode dataset are presented in Figure~\ref{fig:quality2}, where we compare our MOFA-VTON with COTTON~\cite{cotton}, a publicly available method designed to support diverse try-on effects via sparse point control.
As shown in Figure~\ref{fig:quality2}, while COTTON demonstrates some variation in clothing adaptation, these differences are relatively minor. Moreover, it suffers from distortion artifacts, particularly along the clothing boundaries.
In contrast, MOFA-VTON can accurately match more complex and fine-grained hem curves.

\begin{table}[!t]
\centering
\footnotesize
\caption{\textbf{Quantitative comparisons with baseline methods on the VITON-HD dataset.} The background colors red, orange, and yellow in the table represent the best, second-best, and third-best results, respectively.}
\setlength{\tabcolsep}{1.2pt}{
\begin{tabular}{l|cccc|cc}
\toprule
\multicolumn{1}{c}{\multirow{2}{*}{Method}}                  & \multicolumn{4}{c}{Paired}   & \multicolumn{2}{c}{Unpaired}  \\                  
\cmidrule{2-7}     \multicolumn{1}{c}{}             & FID (\(\downarrow\))    & KID (\(\downarrow\))     & SSIM (\(\uparrow\))     & \multicolumn{1}{c}{LPIPS (\(\downarrow\))}   & FID (\(\downarrow\))      & KID (\(\downarrow\))    \\ 
\midrule \midrule
HR-VTON~\cite{hr-vton}                 & 10.01         &  3.63         &  0.8642         &  0.0845             & 12.23          & 3.87       \\
TPD~\cite{tpd}                                                    & 9.49          &  5.01         &  0.8715         &  0.0700             & 12.43          &  5.31      \\
GarDiff~\cite{gardiff}                                            & 8.94          &  4.19         &  0.8503         &  0.0786             & 11.42          &  4.35      \\
DCI-VTON~\cite{dci-vton}                                          &  6.64         &  1.58         &  0.8768         &  0.0740             &  9.58          &  1.91      \\
SD-VITON~\cite{sd-viton}                                          &  6.85         &  \cc{2} 0.98  &  0.8742         &  0.0854             &  9.52          &  1.40      \\
LaDI-VTON~\cite{ladi-vton}                                        &  6.80         &  1.28         &  0.8625         &  0.0803             &  9.38          &  1.66      \\
CAT-DM~\cite{cat-dm}                                              & 6.79          &  2.39         &  0.8677         &  0.0773             & 10.33          &  2.56      \\
IDM-VTON~\cite{idm-vton}                                          &  6.45         &  1.46         &  0.8635         &  0.0700             &  9.37          &  1.58      \\
GP-VTON~\cite{gp-vton}                                            &  \cc{3} 6.41  &  \cc{3} 1.04  &  \cc{3} 0.8839  &  \cc{3} 0.0669      &  \cc{3} 9.34   &  \cc{3} 1.23      \\
StableVITON~\cite{stableviton}                                    &  \cc{2} 6.05  &  1.09         &  \cc{2} 0.8867  &  \cc{1} 0.0605      &  \cc{2} 9.14   &  \cc{2} 1.31      \\
\textbf{MOFA-VTON}                                         &  \cc{1} 5.97  & \cc{1}  0.92  &  \cc{1} 0.8870  &  \cc{2} 0.0632      &  \cc{1} 8.61   &  \cc{1} 1.17 \\ 
\bottomrule
\end{tabular}
\label{tab:quantitative}}
\end{table}
\begin{table}[!t]
\centering
\footnotesize
\caption{\textbf{Quantitative comparisons with baseline methods on the DressCode upper-body (D.C. Upper) dataset.}}
\setlength{\tabcolsep}{1.2pt}{
\begin{tabular}{l|cccc|cc}
\toprule
\multicolumn{1}{c}{\multirow{2}{*}{Method}}                  & \multicolumn{4}{c}{Paired}   & \multicolumn{2}{c}{Unpaired}  \\                  
\cmidrule{2-7}     \multicolumn{1}{c}{}             & FID (\(\downarrow\))    & KID (\(\downarrow\))     & SSIM (\(\uparrow\))     & \multicolumn{1}{c}{LPIPS (\(\downarrow\))}   & FID (\(\downarrow\))      & KID (\(\downarrow\))    \\ 
\midrule \midrule
COTTON~\cite{cotton}                  &  16.54          & 4.67            &  0.8871            &  0.0981          &  17.29         & 5.47       \\
LaDI-VTON~\cite{ladi-vton}                                        &  10.75          & 2.50            &  0.9170            &  0.0472          &  13.99         & 3.03       \\
CAT-DM~\cite{cat-dm}                                              &  8.21           & \cc{3} 1.02     &  0.9283            &  0.0395          &  \cc{3} 12.19  & 1.99       \\
GP-VTON~\cite{gp-vton}                                            & \cc{3} 7.60     & \cc{2} 0.85     &  \cc{2} 0.9434     &  \cc{3} 0.0323   &  12.48         & \cc{2} 1.35       \\
IDM-VTON~\cite{idm-vton}                                          & \cc{2} 7.36     & 1.09            &  \cc{3} 0.9362     &  \cc{1} 0.0291   &  \cc{2} 11.73  & \cc{3} 1.68       \\
\textbf{MOFA-VTON}                                         & \cc{1} 6.41     & \cc{1} 0.72     &  \cc{1} 0.9452     &  \cc{2} 0.0316   &  \cc{1} 9.17   & \cc{1} 1.06 \\
\bottomrule
\end{tabular}
\label{tab:quantitative2}}
\end{table}

\begin{figure*}
    \centering
    \includegraphics[width=1\linewidth]{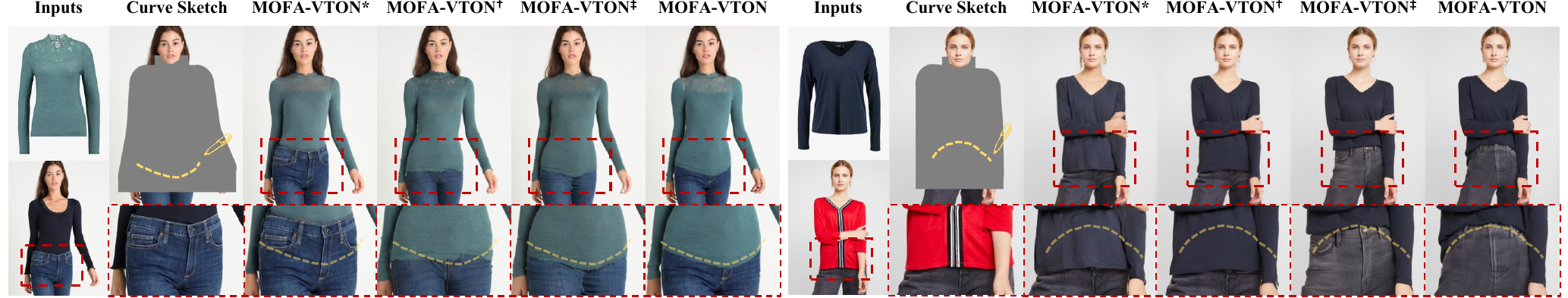}
    \caption{\textbf{Qualitative results of ablation studies.} The bottom of each column presents a zoomed-in view.}
    \label{fig:ab}
\end{figure*}

\subsection{Quantitative Results}
\label{quan}
We conduct quantitative experiments in both paired and unpaired data settings. The paired setting represents that the target clothing matches the clothing originally worn by the person, while the unpaired setting involves a person wearing different clothing after try-on. 
For evaluation, we use Structural Similarity (SSIM)~\cite{ssim} and Learned Perceptual Image Patch Similarity (LPIPS)~\cite{lpips} in the paired setting, along with Fréchet Inception Distance (FID)~\cite{fid} and Kernel Inception Distance (KID)~\cite{kid} in both paired and unpaired settings.
To ensure a fair comparison for try-on quality, we perform the metric evaluation using try-on results under the default effect settings, where the clothing adaptation remains consistent with the original image, similar to other methods.
Table~\ref{tab:quantitative} and Table~\ref{tab:quantitative2} present comparisons on the VITON-HD and DressCode upper-body (D.C. Upper) datasets, respectively. The results indicate that MOFA-VTON outperforms all baselines in most metrics, with only LPIPS achieving second-best performance, showcasing its ability to deliver high-quality and realistic try-on results.

\begin{table}[!t]
\centering
\footnotesize
\caption{\textbf{Quantitative results of ablation studies.} The variants respectively ablate the dual-region mask, layout adjustment blocks, and the feature expansion and filtering.}
\setlength{\tabcolsep}{1.2pt}{
\begin{tabular}{l|cccc|cc}
\toprule
\multicolumn{1}{c}{\multirow{2}{*}{Method}}        & \multicolumn{4}{c}{Paired}    &\multicolumn{2}{c}{Unpaired} \\ 
\cmidrule{2-7}  \multicolumn{1}{c}{}              & FID ($\downarrow$)     & KID ($\downarrow$)       & SSIM ($\uparrow$)    & \multicolumn{1}{c}{LPIPS ($\downarrow$)}    & FID ($\downarrow$)    & KID ($\downarrow$)  \\ 
\midrule \midrule        
MOFA-VTON$^{\ast}$           & 7.79          & 1.59          & 0.8697         & 0.0826            & 11.10         & 2.50          \\
MOFA-VTON$^{\dagger}$        & 6.52          & 1.23          & 0.8717         & 0.0661            & 9.06          & 1.80          \\
MOFA-VTON$^{\ddagger}$       & 6.10          & 1.01          & 0.8818         & 0.0649            & 8.78          & 1.19          \\ \midrule
MOFA-VTON                    & \textbf{5.97} & \textbf{0.92} & \textbf{0.8870} & \textbf{0.0632}  & \textbf{8.61} & \textbf{1.17}  \\ 
\bottomrule
\end{tabular}}
\label{tab:ablation}
\end{table} 

\subsection{User Study}
We conduct a user study involving 35 volunteers to evaluate the visual quality and diversity of our method, where MOFA-VTON is compared against each baseline in an A/B evaluation.
For each comparison, we randomly select 200 clothing images and 200 person images from the test set of the VITON-HD dataset. Try-on results are then generated with both MOFA-VTON and the baseline method, forming 200 paired image groups.
Volunteers are instructed to assess each pair of try-on results in terms of both quality and diversity, and to select the one they prefer. The diversity is further divided into four items: fitness, usability, functionality, and fineness.
As shown in Figure~\ref{fig:user_study}, the results clearly show that MOFA-VTON holds an advantage in the vast majority of A/B comparison items.

\subsection{Ablation Studies}
\noindent\textbf{Dual-region Mask.}
To assess the impact of the dual-region mask as layout guidance, we design a variant MOFA-VTON$^{\ast}$ that ablates the dual-region mask and replaces it with a conventional clothing-agnostic mask. Due to the lack of a separate region representation, the subsequent region encoder and layout adjustment blocks are also removed.
We first compare this variant and our full model in Figure \ref{fig:ab}. 
It is evident that MOFA-VTON$^{\ast}$ can only fit the target clothing onto the person in a fixed manner identical to the original image.
Quantitative results of MOFA-VTON$^{\ast}$ and MOFA-VTON are reported in Table \ref{tab:ablation}, which shows that the full model notably outperforms the variant across all metrics.

\noindent\textbf{Layout Adjustment Block.}
We design a variant MOFA-VTON$^{\dagger}$ to verify the effect of layout adjustment blocks.
For this variant, we remove all the layout adjustment blocks placed in Adapt-Net, therefore the region encoder providing input features for them also becomes ineffective, the dual-region mask is fed into the model solely by concatenating it with the noised image and DensePose map.
MOFA-VTON$^{\dagger}$ and MOFA-VTON are quantitatively compared in Table \ref{tab:ablation}, while Figure \ref{fig:ab} illustrates their differences in visual effects. 
It can be seen that although MOFA-VTON$^{\dagger}$ can slightly adjust the clothing layout, it tends to fail when encountering complex poses like the second example.

\noindent\textbf{Feature Expansion and Filtering.}
In our full model, we incorporate transposed convolutional layers during the mask-guided layout adjustment to extend the influence of region features while applying mask operations to filter out irrelevant information.
To assess their effectiveness, we introduce MOFA-VTON$^{\ddagger}$, which removes all transposed convolution and subsequent mask operations.
The quantitative metrics for MOFA-VTON$^{\ddagger}$ are reported in Table~\ref{tab:ablation}, and the qualitative comparison with the full model is shown in Figure~\ref{fig:ab}.
It can be observed that MOFA-VTON$^{\ddagger}$ effectively adjusts clothing adaptation to roughly match the user-drawn curves. However, compared to the full model, it still lacks sufficient refinement in the details of the clothing hem.
\section{Conclusion}
\label{conclusion}
In this paper, we propose MOFA-VTON, a novel virtual try-on method to delve into more fashion possibilities with diverse and fine-grained adaptations. 
Through the carefully designed dual-region mask and layout adjustment blocks, MOFA-VTON enables the target clothing to be adjusted on the body according to a user-drawn curve, achieving highly flexible and diverse try-on effects.
Extensive experiments demonstrate that MOFA-VTON outperforms existing methods for virtual try-on in terms of quality, diversity, and fine-grained control.
Overall, MOFA-VTON significantly enhances the virtual try-on experience, making it more closely aligned with reality and offering more fashion possibilities.

\section{Acknowledgement} 
This work was supported by the National Natural Science Foundation of China (Grant Nos. U2574212, 62272134, 62572152, and 62502124) and the Natural Science Foundation of Shandong Province (Grant Nos. ZR2024QF065 and ZR2025MS995).

{
    \small
    \bibliographystyle{ieeenat_fullname}
    \bibliography{main}
}
 
\newpage
\appendix
\label{appendix}

\section{Implementation Details}
In our experiments, we initialize the weights of Cloth-Net and Adapt-Net with the pre-trained weights of Paint-by-Example~\cite{paint_by_example}, and the weights of the CLIP~\cite{clip} model are taken from the version of ViT-L/14.
Our model is trained using the AdamW optimizer~\cite{adamW} with a learning rate of 1e-5. The training is conducted on paired images with a resolution of 512$\times$384, and we adopt a batch size of 8 throughout the training process. For inference, the whole virtual try-on pipeline can be executed in approximately 5.7 seconds when running on a single NVIDIA A100 GPU.

\section{User Study Details}
\label{appendixB}
To provide a more comprehensive evaluation of our proposed MOFA-VTON, we conduct a detailed user study focusing on diversity. In this study, diversity is further divided into four items: fitness, usability, functionality, and fineness.
Specifically, fitness is designed as a purely subjective metric to assess users’ general preference for diverse try-on results without being influenced by other factors, which measures whether the result overall meets user expectations. 
Usability evaluates the ease of operation and user interaction of each method. Therefore, no visual results are shown for this criterion, where participants are presented with textual descriptions of the interaction workflows and asked to select the method they considered more convenient and intuitive. 
Functionality evaluates whether additional or extended functions are achieved, where we conduct a region-wise evaluation by asking participants to score whether each method supports controllable try-on in four body regions (waist, arms, torso, and legs). 
Fineness assesses the level of fine-grained control achieved, where we explicitly visualize the interaction traces on the try-on results (e.g., control points before and after manipulation for COTTON~\cite{cotton}, input curves for MOFA-VTON), allowing participants to judge the precision and granularity of layout control based on these visual cues.

\section{Limitations}
\begin{figure}
    \centering
    \includegraphics[width=1.0\linewidth]{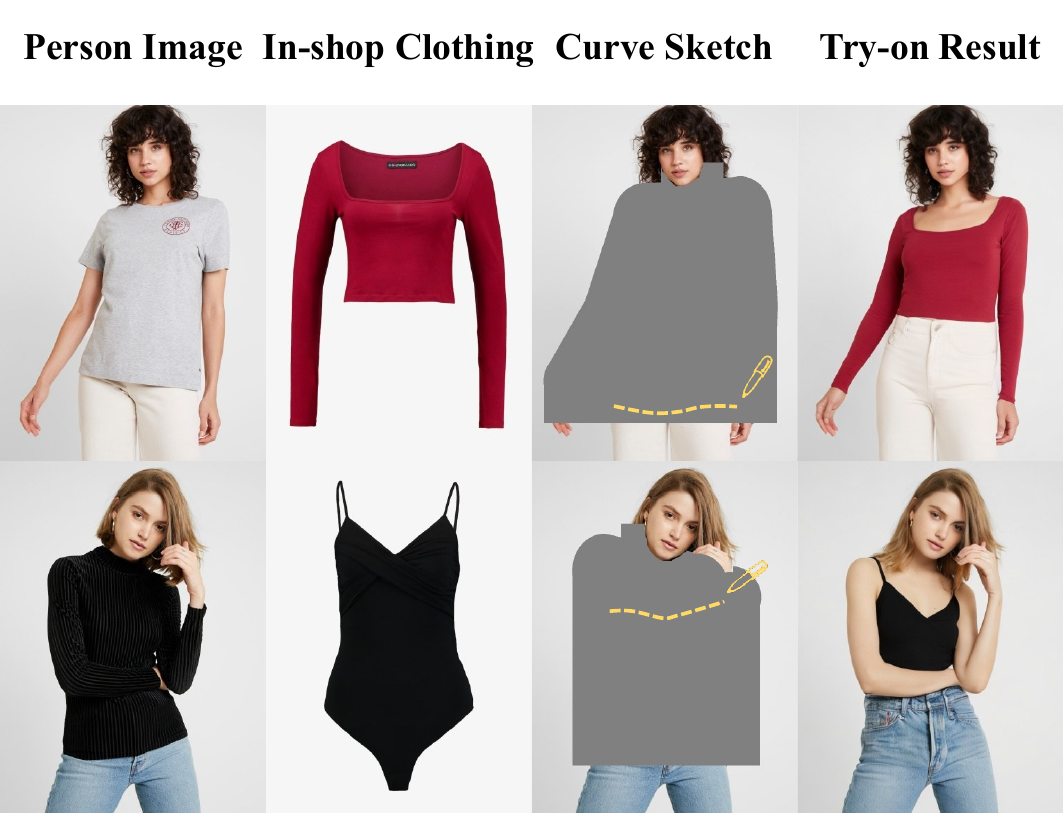}
    \caption{MOFA-VTON struggles to accurately adjust clothing adaptation based on curves positioned at unrealistic locations.}
    \label{fig:limitation}
\end{figure}

Since the training data for our model is entirely collected from real‑world settings under normal and practical conditions, MOFA-VTON learns to predict clothing adaptations that conform to realistic scenarios. 
Therefore, when presented with unusual or extreme input curves, the generated output may not fully match the intended layouts.
As illustrated in Figure~\ref{fig:limitation}, for certain clothing categories such as short-length tops or bodysuits, input curves positioned too low or too high provide ineffective guidance for consistent clothing adaptation.

\begin{figure}
    \centering
    \includegraphics[width=1.0\linewidth]{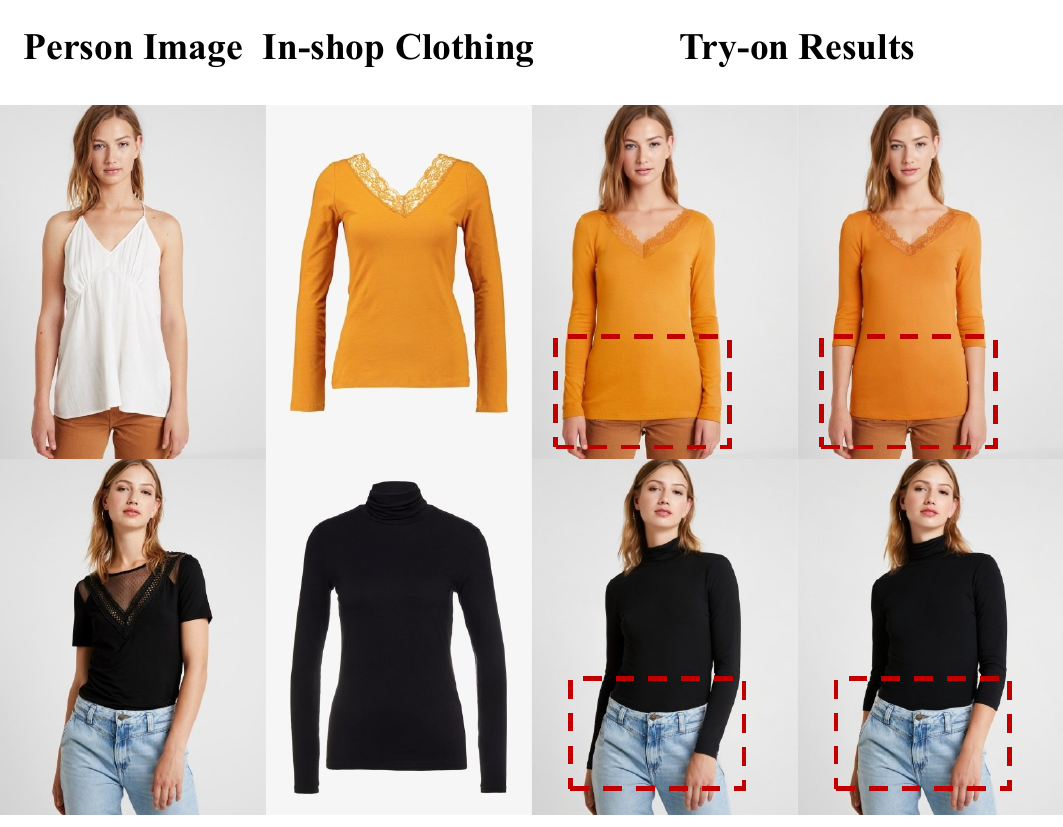}
    \caption{MOFA-VTON enables adjustment of sleeve length.}
    \label{fig:limb}
\end{figure}

\begin{figure*}
    \centering
    \includegraphics[width=1.0\linewidth]{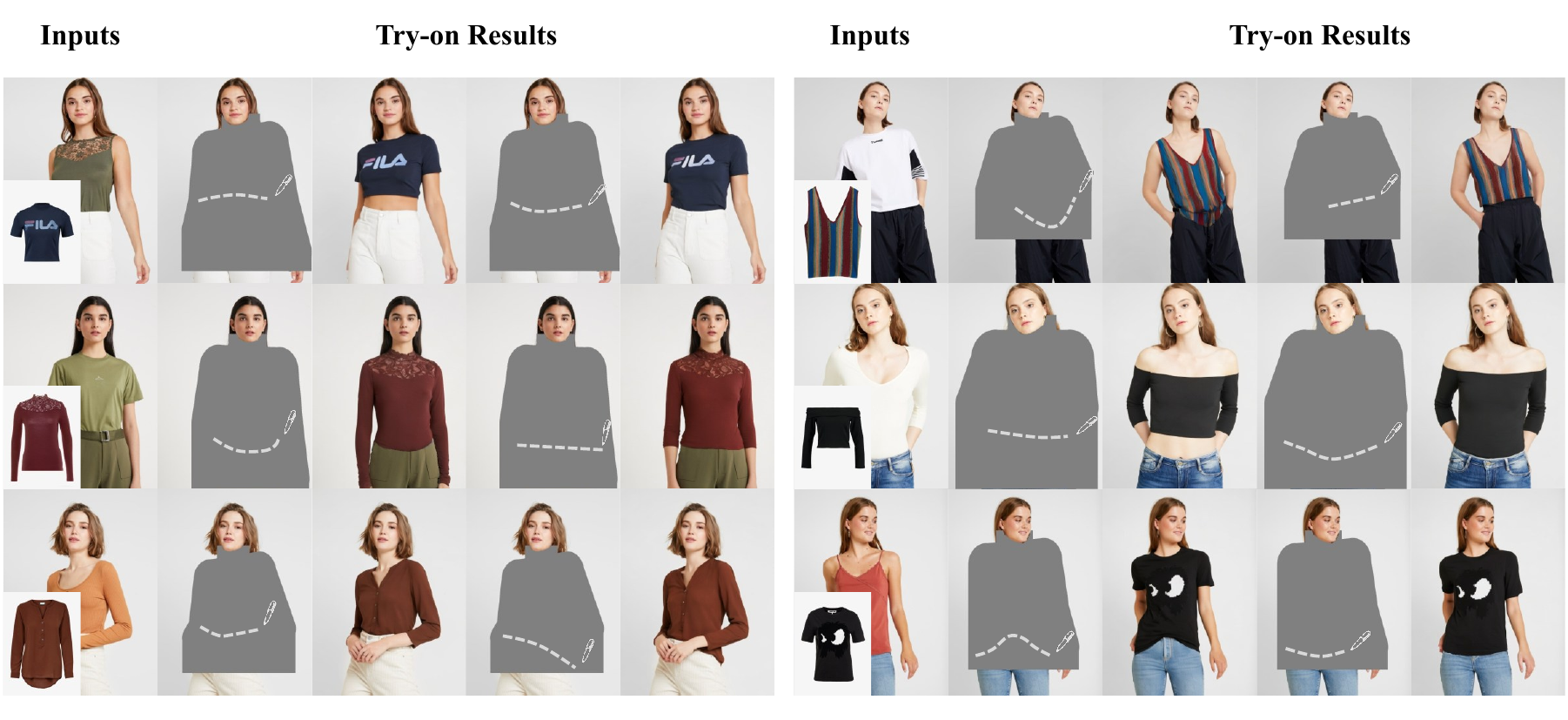}
    \caption{More qualitative results generated by MOFA-VTON.}
    \label{fig:more_res}
\end{figure*}

\begin{table*}[!t]
\centering
\caption{Quantitative results of ablation studies on the CLIP text embedding.}
\setlength{\tabcolsep}{8pt}{
\begin{tabular}{l|cccc|cc}
\toprule
\multicolumn{1}{c}{\multirow{2}{*}{Method}}        & \multicolumn{4}{c}{Paired}    &\multicolumn{2}{c}{Unpaired} \\ 
\cmidrule{2-7}  \multicolumn{1}{c}{}              & FID ($\downarrow$)     & KID ($\downarrow$)       & SSIM ($\uparrow$)    & \multicolumn{1}{c}{LPIPS ($\downarrow$)}    & FID ($\downarrow$)    & KID ($\downarrow$)  \\ 
\midrule \midrule        					
MOFA-VTON w/o CLIP text embedding          & 6.02          & 0.98          & 0.8861         & 0.0633            & 8.77         & 1.20          \\ \midrule
MOFA-VTON                                           & \textbf{5.97} & \textbf{0.92} & \textbf{0.8870} & \textbf{0.0632}  & \textbf{8.61} & \textbf{1.17}  \\ 
\bottomrule
\end{tabular}}
\label{tab:ablation_text}
\end{table*}

\section{More Results}
\label{appendixD}

\noindent\textbf{Additional Ablation Study.} 
In the design of the region encoder, we regard the CLIP text embedding as a complementary enhancement, which is introduced as a high-level semantic prior to guide the layout prediction process. 
By incorporating multi-modal inputs, the network is able to more effectively capture spatial arrangements and clothing semantics.
To assess its contribution, we construct an additional variant that ablates the CLIP text embedding from the region encoder.
As shown in Table~\ref{tab:ablation_text}, there is a slight performance degradation when removing the text embedding, and we also observe that the training process became less stable, especially in the early stages. These results indicate that the text encoder benefits both overall performance and convergence stability.

\begin{figure*}
    \centering
    \includegraphics[width=1.0\linewidth]{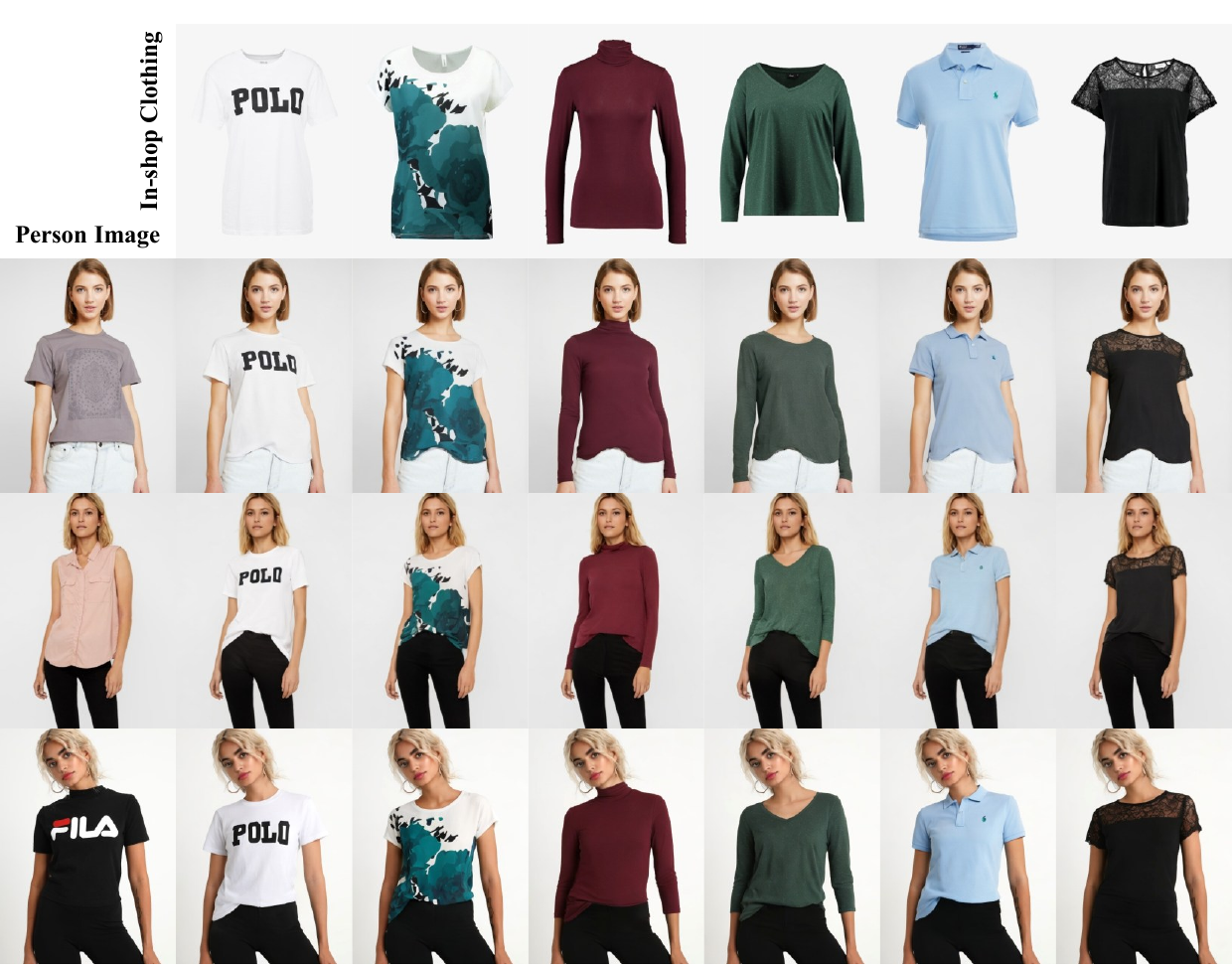}
    \caption{Each person image is paired with various clothing options to generate try-on results with consistent clothing adaptations.}
    \label{fig:cross}
\end{figure*}

\noindent\textbf{Additional Visual Displays.} 
To further demonstrate the fine-grained control capability of our MOFA-VTON, we pair each person image with various clothing options and assign a fixed hand-drawn curve to each group, generating multiple results with consistent clothing adaptations within the same group.
As shown in Figure~\ref{fig:cross}, MOFA-VTON effectively fits the target clothing of varying appearances to the human body and accurately adapts them according to the input curves, further highlighting its strong performance and robustness.
In addition, more qualitative results are also provided in Figure~\ref{fig:more_res}.

\begin{figure*}
    \centering
    \includegraphics[width=1.0\linewidth]{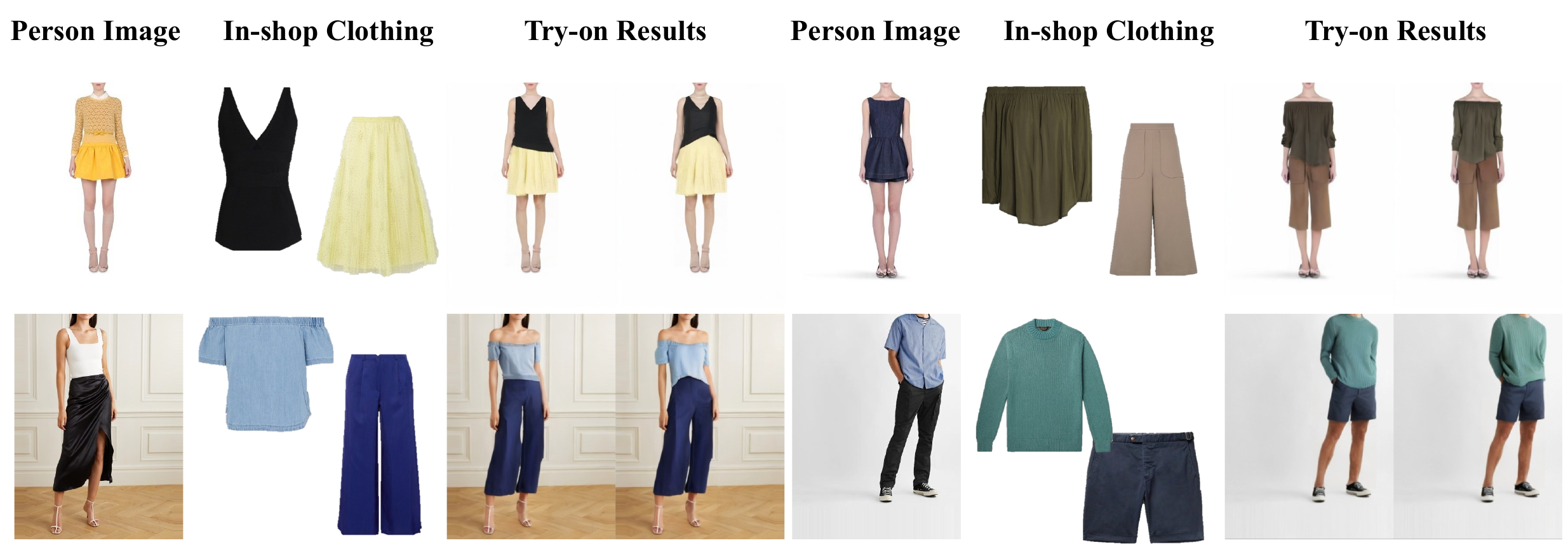}
    \caption{Multi-item try-on results generated by MOFA-VTON.}
    \label{fig:lower_body}
\end{figure*}

\noindent\textbf{Additional Function.} 
MOFA-VTON also supports virtual try-on for other types of clothing such as pants and skirts. This can be achieved by modifying the mask expansion step of the mask construction strategy, where the expansion is performed from the highest point along the y-axis instead of the lowest. Therefore, we conduct multi-item try-on experiments, which are presented in Figure~\ref{fig:lower_body}. These results demonstrate the generalization capability and adaptability of MOFA-VTON across diverse clothing categories.
Moreover, although MOFA-VTON primarily focuses on the interaction between tops and bottoms, we are pleasantly surprised to discover that by slightly modifying the arm-occlusion mask, MOFA-VTON can further support adjusting the sleeve length of clothing, as shown in Figure~\ref{fig:limb}.
We speculate that this effect arises from the model learning strong control over clothing boundaries in the training phase, enabling it to generate realistic details (e.g., wrinkles) around these regions, and this capability generalizes to other regions as well.

\end{document}